\documentclass[sigconf]{aamas}  %

\renewcommand\footnotetextcopyrightpermission[1]{} %
\usepackage{booktabs}

\usepackage{flushend}
\setcopyright{ifaamas}  %
\acmDOI{doi}  %
\acmISBN{}  %
\acmConference[AAMAS'20]{Proc.\@ of the 19th International Conference on Autonomous Agents and Multiagent Systems (AAMAS 2020), B.~An, N.~Yorke-Smith, A.~El~Fallah~Seghrouchni, G.~Sukthankar (eds.)}{May 2020}{Auckland, New Zealand}  %
\acmYear{2020}  %
\copyrightyear{2020}  %
\acmPrice{}  %

\usepackage{microtype}
\usepackage{graphicx}
\usepackage{float}
\usepackage{booktabs} %
\usepackage{algorithm}
\usepackage{algorithmic}

\usepackage{caption}
\usepackage{subcaption}

\usepackage{amsmath,amssymb,amsthm}

\newtheorem{assumption}{Assumption}

\newcommand{\argmax}{\operatorname*{argmax}} %

\newcommand{\kl}{\operatorname*{KL}}
\newcommand{\clip}{\operatorname{clip}}
\newcommand{\relu}{\operatorname{relu}}

\newcommand{\E}{\mathbb{E}}

\newcommand{\g}{\mathcal{G}}

\newcommand{\R}{\mathbb{R}}

\begin{document}

\title{Modified Actor-Critics}  %

\author{Erinc Merdivan}
\affiliation{
\institution{Austrian Institute of Technology}
}

\author{Sten Hanke}
\affiliation{
\institution{FH Joanneum}
}

\author{Matthieu Geist}
\affiliation{
\institution{Google Research, Brain Team}
}

\begin{abstract}

Recent successful deep reinforcement learning algorithms, such as Trust Region Policy Optimization (TRPO) or Proximal Policy Optimization (PPO), are fundamentally variations of conservative policy iteration (CPI). These algorithms iterate policy evaluation followed by a softened policy improvement step. As so, they are naturally on-policy.
In this paper, we propose to combine (any kind of) soft greediness with Modified Policy Iteration (MPI). The proposed abstract framework applies repeatedly: (i) a partial policy evaluation step that allows off-policy learning and (ii) any softened greedy step.
Our contribution can be seen as a new generic tool for the deep reinforcement learning toolbox.
As a proof of concept, we instantiate this framework with the PPO greediness. Comparison to the original PPO shows that our algorithm is much more sample efficient. We also show that it is competitive with the state-of-art off-policy algorithm Soft Actor Critic (SAC).
\end{abstract}

\maketitle

\section{Introduction}

Many deep reinforcement learning (RL) algorithms are based on approximate dynamic programming. For example, the celebrated DQN~\citep{mnih2015human} is based on approximate value iteration. As a pure critic approach, it can only deal with finite action spaces. A more versatile approach, that allows handling both discrete and continuous action spaces, consists in considering actor-critic architectures, where both the value function and the policy are represented. Most of such recent approaches are either variations of policy gradient~\citep{lillicrap2015continuous,wang2016sample,mnih2016asynchronous}, inspired by conservative policy iteration~\citep{schulman2015trust,schulman2017proximal,wu2017scalable}, or make use of entropy regularization~\citep{haarnoja2017reinforcement,haarnoja2018soft,haarnoja2018softb}.

If approximate policy iteration has already been the building block of actor-critics in the past~\citep{gabillon2011classification}, it has not been considered with deep learning approximators, as far as we know. We assume that this is due to the fact that the greedy operator is unstable (much like gradient descent with too big step sizes). A clever way to address this issue has been introduced by~\citet{kakade2002approximately} with Conservative Policy Iteration (CPI). Instead of taking the greedy policy, the new policy is a stochastic mixture of the current one and of the greedy one. This softens greediness and stabilizes learning.

With this classical approach, the current policy is a stochastic mixture of all past policies, which is not very practical. The core idea of CPI has been adapted in the deep RL literature by modifying how the greediness is softened. For example, Trust Region Policy Optimization (TRPO)~\citep{schulman2015trust} or  Actor-Critic using Kronecker-factored Trust Region (ACKTR)~\citep{wu2017scalable} add a constraint on the greedy step, imposing that the average Kullback-Leibler (KL) divergence between consecutive policies is below a given threshold, and Proximal Policy Optimization (PPO)~\citep{schulman2017proximal} modifies the greedy step with a clipping loss that forces the ratio of action probabilities of consecutive policies to remain close to 1. To some extent, even policy gradient approaches can be seen as such, as following the policy gradient should provide a softened improvement (see also~\citep{scherrer2014local} for a connection between CPI and policy gradient). Other approaches consider an entropy penalty~\citep{haarnoja2017reinforcement,haarnoja2018soft,haarnoja2018softb}, which effect is also to soften greediness (but can also modify the evaluation step).

In this paper, we will call generally ``Softened Policy Iteration'' (SPI) any approach that combines policy evaluation with a softened greedy step. As they require policy evaluation, these approaches are naturally on-policy. In classical dynamic programming, Modified Policy Iteration (MPI)~\citep{puterman1978modified} replaces the full evaluation of the policy by a partial evaluation. This idea has been extended to the approximate setting (Approximate MPI, or AMPI~\citep{scherrer2012approximate}), but never with deep learning approximators, as far as we know. This is probably due to the instability of the greedy step.

Yet, a partial evaluation presents some interest, compared to a full policy evaluation. It allows for an easier extension to off-policy learning by making use of Temporal Difference (TD) learning instead of using rollouts. It also draws a bridge between value and policy iterations (because MPI has these two algorithms as special cases). In this work, we propose an abstract actor-critic framework that brings together MPI and SPI, by mixing the partial evaluation of MPI with the softened greediness of SPI. We name the resulting approach Modified Soft Policy Iteration (MoSoPI).

To justify this approach, we show that MoSoPI converges in the ideal case (no approximation error), and discuss briefly in what cases it can converges to an optimal policy. As a proof of concept of this general simple idea, we instantiate it with the PPO greediness, and compare it to the original PPO on a set of continuous control tasks~\citep{todorov2012mujoco}. The only difference between both algorithms is  the way state(-action) value functions are (partially) estimated, yet it allows gaining a lot regarding sample efficiency. To be complete, we'll also compare this modified PPO to a state of the art off-policy actor-critic, Soft Actor-Critic (SAC)~\citep{haarnoja2018soft}. It is often competitive with it, while being usually more sample efficient.%

\section{Background}

A Markov Decision Process (MDP) is a tuple $\{S,A,P,r,\gamma\}$, with $S$ the state space, $A$ the action space, $P$ the transition kernel ($P(s'|s,a)$ denotes the probability to go from $s$ to $s'$ under action $a$), $r\in\mathbb{R}^{S\times A}$ the reward function and $\gamma\in(0,1)$ the discount factor. A (stochastic) policy $\pi$ is a mapping from states to distribution of actions ($\pi(a|s)$ denotes the probability of choosing $a$ in $s$). The quality of a policy is quantified by the value function,
\begin{equation}
    v_\pi(s) = \E_\pi\left[\sum_{t\geq 0} \gamma^t r(s_t,a_t)|s_0=s\right],
\end{equation}
where $\E_\pi$ denotes the expectation respectively to the trajectories sampled by the policy $\pi$ and the dynamics $P$.

Write $T_\pi$ the Bellman operator, defined for any $v\in\mathbb{R}^S$ as
\begin{equation}
    \forall s\in S,\quad [T_\pi v](s) = \E_{a\sim\pi(.|s)}[r(s,a) + \gamma v(s')].
\end{equation}
The value function $v_\pi$ is the unique fixed point of the operator $T_\pi$. The aim of RL is to maximize either the value function for each state or an average value function. To do so, the notion of Bellman optimality operator is useful:%
\begin{equation}
    \forall v\in\mathbb{R}^S, \quad T v = \max_\pi T_\pi v.
\end{equation}
The optimal value function $v_*$ is the unique fixed point of $T$. The notion of greedy operator can be derived from $T$. We say that $\pi$ is greedy respectively to $v\in\mathbb{R}^S$ (that is not necessarily a value function) if
\begin{equation}
    \pi\in\g(v) \Leftrightarrow T v = T_\pi v.
\end{equation}

The value function might not be convenient from a practical viewpoint, as applying the operators $T$ and $\g$ requires knowing the dynamics. To alleviate this issue, a classical approach is to consider a $Q$-function, that adds a degree of freedom on the first action to be chosen, %
\begin{equation}
    Q_\pi(s,a) = \E_\pi\left[\sum_{t\geq 0} \gamma^t r(s_t,a_t)|s_0=s, a_0=a\right].
\end{equation}
Similarly to the value function, we can define the associated $T_\pi$, $T$ and $\g$ operators. Value and $Q$-functions are linked by $v_\pi(s) = \E_{a\sim\pi(.|s)}[Q_\pi(s,a)]$, and the advantage function is defined as the state-wise centered $Q$-function, $A_\pi(s,a) = Q_\pi(s,a) - v_\pi(s)$.

\section{Modified Softened Policy Iteration}

In this section, we  present the abstract variations of policy iteration that lead to MoSoPI, as well as briefly how they can be transformed into practical algorithms. We also justify MoSoPI by showing its convergence in an ideal case.

\subsection{Policy Iteration}
\label{subsec:pi}

Policy iteration (PI) alternates policy improvement and policy evaluation:
\begin{equation}
    \begin{cases}
        \pi_{k+1} = \g(v_k)
        \\
        v_{k+1} = v_{\pi_{k+1}}
    \end{cases}.
    \label{eq:pi}
\end{equation}
In the exact case, everything can be computed analytically (given finite and small enough state and action spaces), and this PI scheme will converge in finite time. In an approximate setting, one has to approximate both the value function and the policy (possibly implicitly), and to learn them from samples.

We start by discussing the approximation of policy evaluation. First, as explained before, it is more convenient to work with $Q$-functions. Let $Q_\theta$ be a parameterized $Q$-function, $Q_\pi$ can be estimated using rollouts. Write generally $\hat{\E}$ for an empirical estimation, assume that a set of state-action couples $(s_i,a_i)_{1\leq i\leq n}$ is available, and that we can simulate the return $R_i$ (the cumulative discounted reward from a rollout starting in $(s_i,a_i)$ and following the policy afterwards), then the $Q$-function can be estimated by minimizing
\begin{equation}
    J(\theta) = \hat{\E}\left[(R_i - Q_\theta(s_i,a_i))^2\right].
\end{equation}
There exist approaches for estimating $Q$-functions directly from transitions, such as LSTD~\citep{bradtke1996linear}, but they usually assume a linear parameterization.

If the action space is finite, the greedy policy can be deduced from the estimated $Q$-function $\hat{Q}_k$:
\begin{equation}
    \pi_{k+1}(a|s) = \begin{cases} 1 &\text{ if } a = \argmax_b \hat{Q}_k(s,b) \\ 0 &\text{ else } \end{cases}.
\end{equation}
Generally, one can also adopt a parameterized policy $\pi_w$ and solve the greedy step as maximizing the following optimization problem:
\begin{equation}
    J(w) = \hat{\E}\left[ \E_{a\sim\pi_w(.|s_i)} [\hat{Q}_k(s_i,a)]\right].
    \label{eq:approx-greedy}
\end{equation}
Notice that this would correspond to solving $\E_{s\sim\mu}[[T_{\pi_w} v](s)]$ for some distribution $\mu$ instead of the greedy step in~\eqref{eq:pi}.
Adding a state-dependant baseline to $\hat{Q}_k$ does not change the minimizer, and one consider usually an estimated advantage function $\hat{A}_k$ to reduce the variance of the gradient. With discrete actions, this corresponds to a cost-sensitive multi-class classification problem~\cite{gabillon2011classification}.

\subsection{Softened Policy Iteration}
\label{subsec:cpi}

The greedy step can be unstable in an approximate setting. To alleviate this problem, \citet{kakade2002approximately} proposed to soften it by mixing the greedy policy with the current one. Let $\alpha_k\in(0,1)$, the greedy step $\pi_{k+1} \in \g(v_k)$ is replaced by
\begin{equation}
    \pi_{k+1} = (1-\alpha_k) \pi_k + \alpha_k \g(v_k).
\end{equation}
This comes with a monotonic improvement guarantee, given a small enough $\alpha_k$. However, it is not very practical, as the new policy is a mixture of all previous policies.

To alleviate this problem, \citet{schulman2015trust} proposed to soften the greediness with a KL penalty between consecutive policies, that leads to minimize:
\begin{align}
    \hat{\E}\left[ \E_{a\sim\pi_w(.|s_i)} [\hat{Q}_k(s_i,a)]\right] 
    \text{ s.t. } 
    \hat{\E}[\kl(\pi_w(.|s_i)||\pi_k(.|s_i))]\leq \epsilon. 
\end{align}
Other approaches are possible. For example, PPO combines the approximate greedy step~\eqref{eq:approx-greedy} with importance sampling and a clipping of the ratio of probabilities:
\begin{equation}
    J(w) = \hat{\E}\left[ \E_{a\sim\pi_k(.|s_i)} \left[\clip_\epsilon\left(\frac{\pi_w(a|s_i)}{\pi_k(a|s_i)}\hat{A}_k(s_i,a)\right)\right]\right].
    \label{eq:ppo-greedy-step}
\end{equation}
The $\clip_\epsilon$ operator saturates the ratio of probabilities when it deviates too from 1 (at $1+\epsilon$ if the advantage is positive, $1-\epsilon$ else), without it it would be equivalent to~\eqref{eq:approx-greedy}.

In this work, we call SPI any policy iteration combined with a softened greedy step, that we frame as satisfying $T_{\pi_{k+1}} v_k\geq T_{\pi_{k}} v_k$ (so, we ask the policy to provide some improvement, without being necessarily the greedy one). In that sense, even a policy gradient step can be seen as softened greediness.

\subsection{Modified Policy Iteration}
\label{subsec:mpi}

If SPI modifies the greedy step, MPI~\citep{puterman1978modified} modifies the evaluation step. The operator $T_{\pi_k}$ being a contraction, we can write $v_{\pi_k} = (T_{\pi_k})^\infty v$, for any $v\in\mathbb{R}^S$, so notably for $v=v_{k-1}$. MPI does partial evaluation by iterating the operator a finite number of times. Let $m\geq 1$, MPI iterates
\begin{equation}
    \begin{cases}
        \pi_{k+1} = \g(v_k)
        \\
        v_{k+1} = (T_{\pi_{k+1}})^m v_{k}
    \end{cases}.
    \label{eq:mpi}
\end{equation}
For $m=\infty$, we retrieve PI, and for $m=1$ we retrieve value iteration (VI): as $T_{\g(v)} v = T v$, with $m=1$ it reduces to $v_{k+1} = T v_k$.

We have that 
\begin{equation}
    (T_\pi)^m v = \E_\pi\left[\sum_{t=0}^{m-1}\gamma^t r(s_t,a_t) + \gamma^m v(s_m)|s_0=s\right].
\end{equation}
This suggests two ways of estimating a value function (or next, directly a $Q$-function). First, consider the case $m=1$ and a parameterized $Q$-function. The classical approach consists in solving the following regression problem:
\begin{align}
    J(\theta) &= \hat{\E}\left[\left(y_i - Q_\theta(s_i,a_i)\right)^2\right]
    \label{eq:td-loss}
    \\
    \text{ with }
    y_i &= r_i + \gamma \E_{a'\sim \pi_{k+1}(.|s')}[\hat{Q}_k(s',a')].
\end{align}
With $m > 1$, a solution is to perform an $m$-step rollout (using $\pi_{k+1}$) and to replace $y_i$ in Eq.~\eqref{eq:td-loss} by
\begin{equation}
    y_i^m = \sum_{t=0}^{m-1} \gamma^t r_{i+t} + \gamma^m \E_{a'\sim \pi_{k+1}(.|s_{t+m})}[\hat{Q}_k(s_{t+m},a')].
\end{equation}
This can be corrected for off-policy learning, using for example importance sampling or Retrace~\citep{munos2016safe}.

Another approach is to solve $m$ times the regression problem of Eq.~\eqref{eq:td-loss}, replacing $\hat{Q}_k$ by the newly computed $Q_\theta$ after each regression but keeping the policy $\pi_{k+1}$ fixed over the $m$ regressions. In other words, solving Eq.~\eqref{eq:td-loss} is one application of an approximate Bellman evaluation operator, and this amounts to applying it $m$ times.
Although using $m$-step returns is pretty standard in deep RL (even if its relation to the classical MPI is rarely acknowledged, as far as we know), the second approach is less usual and has never been experimented, at least in a deep RL context, to the best of our knowledge.

\subsection{Modified Soft Policy Iteration}

MoSoPI simply consists in bringing together a softened policy step of SPI (so, any kind of softened greediness) and the partial evaluation step of MPI:
\begin{equation}
    \begin{cases}
        \text{find } \pi_{k+1} \text{ s.t. } T_{\pi_{k+1}} v_k \geq T_{\pi_{k}} v_k
        \\
        v_{k+1} = (T_{\pi_{k+1}})^m v_{k}
    \end{cases}.
    \label{eq:MoSoPI}
\end{equation}
To get a practical algorithm, one just has to choose a soft greedy step (eg., one of those presented in Sec.~\ref{subsec:cpi}) and to estimate the partial evaluation of the $Q$-function (eg., with one of the approaches depicted in Sec.~\ref{subsec:mpi}). We present in more details such an instantiation in Sec.~\ref{sec:mo-ppo}, that uses the greedy step of PPO and applies $m$ times the approximate Bellman operator for evaluation. However, before this, we show that the algorithmic scheme~\eqref{eq:MoSoPI} converges in the ideal case (no approximation error).

\begin{assumption}[Initialization]
\label{ass:init}
The initial policy $\pi_0$ and value $v_0$ satisfy $T_{\pi_0} v_0 \geq v_0$.
\end{assumption}

This assumption will allow to show monotonicity of values $v_k$ from the beginning. It is a mild assumption. For example, it is satisfied by taking $v_0 = v_{\pi_0}$. Otherwise, if an initial $v_0$ is not satisfying the assumption, subtracting a large enough constant allows satisfying Asm.~\ref{ass:init}. Let $\mathbf{1} \in \R^{S}$ be the vector whose components are all equal to 1. We define $v'_0 = v_0 - c\mathbf{1}$. We have that
\begin{align}
T_{\pi_0} v'_0 &\geq v'_0
\\
\Leftrightarrow T_{\pi_0}(v_0 - c\mathbf{1}) &\geq v_0 - c\mathbf{1}
\\
\Leftrightarrow (1-\gamma) c \mathbf{1} &\geq v_0 - T_{\pi_0} v_0
\\
\Leftarrow c &\geq \frac{\max_{s\in S}\left(v_0[s] - [T_{\pi_0} v_0](s)\right)}{1-\gamma}.
\end{align}
The last equation provides the lower bound for $c$ such that $v'_0$ and $\pi_0$ satisfy Asm.~\ref{ass:init}.

\begin{theorem}[Convergence of MoSoPI]
\label{th:cv}
Under Asm.~\ref{ass:init}, the sequence of value functions $(v_k)_{k\geq 0}$ converges.
\end{theorem}
\begin{proof}
First, we start by showing by induction that for any $k\geq 0$, we have that $T_{\pi_{k+1}} v_k \geq v_k$. For $k=0$, we have by the greedy step that $T_{\pi_1}v_0 \geq T_{\pi_0} v_0$. By Asm.~\ref{ass:init}, we have $T_{\pi_0} v_0 \geq v_0$. Putting this together, this shows that the induction is true for $k=0$: $T_{\pi_1}v_0 \geq v_0$. Now, assume that $T_{\pi_k} v_{k-1} \geq v_{k-1}$. By monotonicity of the Bellman operator, for any $n\geq 1$,
\begin{align}
T_{\pi_k} v_{k-1} &\geq v_{k-1} 
\\
\Rightarrow (T_{\pi_k})^{n+1} v_{k-1} &\geq (T_{\pi_k})^n v_{k-1}
\geq \dots \geq T_{\pi_k} v_{k-1} \geq v_{k-1}.
\label{eq:proof:monotone}
\end{align}
Then, we have
\begin{align}
    T_{\pi_{k+1}} v_k &\geq T_{\pi_k} v_k & &\text{by the greedy step}
    \\
    &= (T_{\pi_k})^{m+1} v_{k-1} & &\text{as }v_k = (T_{\pi_k})^m v_{k-1}
    \\
    &\geq (T_{\pi_k})^{m} v_{k-1} & &\text{by Eq.~\eqref{eq:proof:monotone}}
    \\
    &= v_{k} & &\text{as }v_k = (T_{\pi_k})^m v_{k-1}.
\end{align}
The induction hypothesis is true at any iteration.

Next, we show that the series of values $(v_k)_{k\geq 0}$ is increasing and bounded. First, we have
\begin{align}
    v_{k+1} &= (T_{\pi_{k+1}})^m v_k & &
    \\
    &\geq T_{\pi_{k+1}} v_k & &\text{by Eq.~\eqref{eq:proof:monotone}}
    \\
    &\geq v_k & &\text{by the induction hypothesis}.
\end{align}
By definition of the Bellman optimality operator and using the induction hypothesis, $v_k \leq T_{\pi_{k+1}} v_k \leq T v_k$. By direct induction, this shows that $v_k\leq T v_k \leq (T)^n v_k \leq (T)^\infty v_k = v_*$. So we have
\begin{equation}
    v_k \leq v_{k+1} \leq v_*.
\end{equation}
The series $(v_k)_{k\geq 0}$ being increasing and upper-bounded, it converges to $\bar{v}$ satisfying $\bar{v}\leq v_*$. 
\end{proof}

Notice that if Thm.~\ref{th:cv} shows the convergence, it tells nothing about to what it converges, nor about the rate of convergence. For example, a CPI scheme with a mixing rate $\alpha_k$ decaying too fast towards zero would not converge to the optimal policy, even without approximation. This is the price to pay for a very general notion of softened greediness, we think that it is not possible to provide a better result without specifying more this soft greediness. However, we can rely on the literature to get convergence and rate of convergence for given smoothed greediness. For example, in~\cite{vieillard2019deep}, the convergence of scheme~\eqref{eq:MoSoPI} with the CPI greediness is studied, and in~\cite{geist2019theory} the convergence when greediness is regularized with a Bregman divergence is studied (which encompasses notably TRPO-like policy updates).

\section{Modified Proximal Policy Optimization}
\label{sec:mo-ppo}

As a proof of concept, we show how PPO can be modified using the MoSoPI idea, and call the resulting algorithm Modified PPO (MoPPO). Both algorithms will have the same greedy step, and differ by the way value functions are estimated.

We partially presented PPO in Sec.~\ref{subsec:cpi}. Its greedy step is depicted in Eq.~\eqref{eq:ppo-greedy-step}. The advantage is estimated with the temporal difference error computed with an approximate value function. The value function is estimated as in Eq.~\eqref{eq:td-loss} (with $v_\theta$ instead of $Q_\theta$), that is it corresponds to one application of the approximate Bellman operator. The advantage can be estimated by a temporal difference (TD) error, $\hat{A}(s_i,a_i) = \delta_i = r_i + \gamma \hat{v}(s'_i) - \hat{v}(s_i)$.  \citet{schulman2017proximal} go further and consider an advantage estimated by combining successive TD errors with eligibility traces. Let $I$ be the length of the trajectory, the advantage is estimated as
\begin{equation}
    \hat{A}(s_i,a_i) = \sum_{t=0}^{I-i+1} (\gamma\lambda)^t \delta_{i+t}.
\end{equation}
These are estimated in an on-policy manner. Notice that in practice a value of $\lambda$ close to 1 is chosen, that makes this close to the (full) rollouts we described in Sec.~\ref{subsec:pi}.

MoPPO uses exactly the same greedy step as PPO~\eqref{eq:ppo-greedy-step}, but the partial evaluation step depicted in Sec.~\ref{subsec:mpi}. We use a replay buffer to store gathered transitions, and we evaluate the $Q$-function, in an off-policy manner, by solving $m$ times the regression problem~\eqref{eq:td-loss}. We estimate the advantage of a state-action couple by subtracting a Monte-Carlo empirical average of the state-action values from the estimated $Q$-function. If $\hat{Q}$ has been estimated based on the policy $\pi$, we sample $a_1,\dots,a_{N_\text{pol}}\sim \pi(.|s)$ and we estimate
\begin{equation}
    \hat{A}(s,a) = \hat{Q}(s,a) - \frac{1}{N_\text{pol}} \sum_{j=1}^{N_\text{pol}} \hat{Q}(s,a_j).
    \label{eq:advantage-estimation}
\end{equation}

MoPPO is summarized in Alg.~\ref{alg:moppo}. 
Current and target $Q$- and policy networks are initialized. The algorithm feeds a replay buffer by interacting with the environment. At regular steps, the $Q$-network is updated $m$ times by doing $m$ optimizations of~\eqref{eq:td-loss} with stochastic gradient descent, using the same policy $\pi_w$ during all optimizations, but updating the target network between each optimization. In the case of continuous actions, the expected state-action value of each next state is estimated using Monte Carlo. This is followed by the optimization of the policy by stochastic gradient ascent on~\eqref{eq:ppo-greedy-step}. As the transitions are sampled from the buffer (that is bigger than the update frequency), MoPPO is off-policy. As it only uses (repeatedly) one-step rollouts, it does not require off-policy corrections.

\begin{algorithm}[tbh]
  \caption{MoPPO}
  \label{alg:moppo}
\begin{algorithmic}
    \STATE Init. replay buffer $D$ to capacity $N$
    \STATE Init. $Q$ function with random weights $\theta$ 
    \STATE Init. $Q^{\text{targ}}$ function with random weights $\theta^-$
    \STATE Init. policy function $\pi$ with random weights $w$ 
    \STATE Init. policy function $\pi_{old}$ with random weights $w^-$
    \STATE Set clipping ratio $\epsilon$
    \FOR{$t=1$ {\bfseries to} $\text{max}\_\text{steps}$}
        \STATE Sample action $a_t \sim \pi(s_t;\omega)$
        \STATE Execute $a_t$ and get reward $r_t$ and next state $s_{t+1}$
        \STATE Store transition $(s_t, a_t, r_t, s_{t+1})$ in $D$
        \STATE Set $s_{t+1} = s_t$
        \IF{$t \% \text{train}\_\text{freq}=0$}
            \FOR{$i=1$ {\bfseries to} $m$}
                \FOR{$j=1$ {\bfseries to} $\text{q}\_\text{steps}$}
                    \STATE Sample a minibatch from $D$ and do a gradient step on~\eqref{eq:td-loss} (with $\hat{Q}_k = Q^{\text{targ}}$ and $\pi = \pi_w$)
                \ENDFOR
                \STATE Update $Q^\text{targ} = Q_\theta$ 
            \ENDFOR
            \FOR{$i=1$ {\bfseries to} $\text{pol}\_\text{steps}$}
                  \STATE Sample a minibatch from $D$ and do a gradient step on~\eqref{eq:ppo-greedy-step}, using the advantage as estimated in~\eqref{eq:advantage-estimation} (with $\pi_w$ unchanged, $\hat{\pi}_k = \pi_{old}$ and  $\hat{Q}=Q^\text{targ}$)
            \ENDFOR
            \STATE  Update $\pi_\text{old} = \pi $
        \ENDIF
    \ENDFOR
\end{algorithmic}
\end{algorithm}

\section{Related Works}

Being built upon MPI and SPI, MoSoPI is obviously related to these approaches. However, the combination of the two induces key differences.

For example, MoPPO can be related to AMPI~\citep{scherrer2012approximate} as both use the evaluation of MPI. However, AMPI does this with $m$-step rollouts, and the $Q$-function is learnt on-policy. Moreover, the greedy step is not softened. AMPI has never been combined with neural networks, as far as we know. We hypothesize that it would be unstable, due to the greedy step\footnote{Indeed, setting $\epsilon$ high enough in MoPPO, we would retrieve a variation of AMPI, and it did not worked in our experiments, not reported here.}. Also, it has never been considered practically with continuous actions.
As it uses a softened greedy step, MoPPO can be related to various approaches such as TRPO~\citep{schulman2015trust}, ACKTR~\citep{wu2017scalable} and even more PPO~\citep{schulman2017proximal}, with which the only difference is the evaluation step. Thanks to this, MoPPO is off-policy, contrary to the preceding algorithms.

As an off-policy deep actor-critic, MoPPO can also be related to approaches such as SAC~\cite{haarnoja2018soft}, DDPG~\cite{lillicrap2015continuous} or TD3~\cite{fujimoto2018addressing}. They share the same characteristics (off-policy, actor-critic), but they are derived from different principles. SAC is build upon entropy-regularized value iteration, while DDPG and TD3 are based on the deterministic policy gradient theorem~\citep{silver2014deterministic}. The proposed MoSoPI framework is somehow more general, as it allows considering any softened greedy step (and thus those of the aforementioned approaches).
Notice that these approaches are made off-policy by (somehow implicitly) replacing the full policy evaluation by a single TD backup. This corresponds to setting $m=1$ in our framework (but learning and sample collection are entangled, contrary to our approach).

Our approach can also be linked to others that could be seen as quite different at a first look. For example, consider Maximum a posteriori Policy Optimisation (MPO)~\citep{abdolmaleki2018maximum}. It is derived using the expectation-maximization principle applied to a relative entropy objective. However, looking at the final algorithm, it is a kind of softened policy iteration approach. The greedy step is close to the one of TRPO, except that the resulting policy is computed analytically on a subset of states, and generalized by minimizing a KL divergence between a policy network and this analytical policy. The evaluation is done by applying the approximate Bellman evaluation operator, combined with something close to $m$-step rollouts corrected by Retrace~\citep{munos2016safe} for off-policy learning. As so, it can be (roughly) seen as an instantiation of the proposed general MoSoPI framework.

\section{Experiments}
\label{sec:exp}

In this section, we experiment with MoPPO, the proof of concept of the proposed general MoSoPI framework. Experiments are done on Mujoco tasks~\cite{todorov2012mujoco}, with the OpenAI gym framework~\cite{brockman2016openai}.
We'll discuss the pros and cons of MoPPO, but we think important to highlight that our core contribution is a generic tool for the deep RL toolbox, rather than a new state-of-the-art actor-critic (MoPPO is just one possible, simple and natural, instantiation of MoSoPI).

We explained earlier that the partial evaluation could be done using either $m$-step rollouts or by applying $m$-times the (approximate) Bellman operator. We have presented MoPPO with the second option, but both could be tried. Therefore, in Sec.~\ref{subsec:mstp_vs_mreg}, we compare both approaches. We observe that using $m$-step returns in our setting is very unstable, even with a Retrace-based off-policy correction. We assume that this is due to a too high degree of ``off-policyness''. 

In Sec.~\ref{subsec:effect_parameters}, we study the influence of some parameters of our algorithm. The parameter $m$, that allows going from VI-like to PI-like approaches, is a first natural candidate.
MoPPO separates sample collection and value/policy optimization. As shown in Alg.~\ref{alg:moppo}, we indeed collect $\text{train}\_\text{freq}$ transitions and then train sequentially the $Q$-function ($m$ times) and the policy. This is not that common in deep RL, to separate both processes, so we also experiment the influence of this parameter.%

In Sec.~\ref{subsec:comparison}, we compare MoPPO to its natural competitor, PPO, on a set of tasks. Our simple modification will learn consistently better and faster (at the possible cost of long-term stability), and requires sometimes up to ten time less samples. Both PPO and MoPPO are run by ourselves, using the OpenAI implementation for PPO. To get a better vision of the efficiency of the proposed approach, we  also compare it to a recent state of the art off-policy actor-critic deep RL algorithm, SAC. For this comparison, we used the results provided by the authors for their experiments running on the same environments\footnote{\url{https://sites.google.com/corp/view/soft-actor-critic}} (but not with the same random seeds nor computer architecture). On most benchmarks, MoPPO performs better and/or faster (still at the possible cost of long-term stability).
The algorithms are evaluated using either the approach of~\citet{haarnoja2018soft} or of~\citet{wu2017scalable}. For the first approach, the policy is evaluated every 1000 steps by using the mean action (instead of sampling). For the second, we average the 10 best evaluation scores acquired so far, that every 1000 steps (that requires keeping track of the best past 10 policies).
Results are averaged over 5 seeds\footnote{For SAC, we used the provided results, corresponding to five seeds, but we do not know their values and how they have been chosen. For PPO and MoPPO, we took the best 5 seeds over 8 seeds, of values evenly spaced between 1000 and 8000. Without this, the results would be a little bit less stable, but it does not change the overall conclusion.}.

In Sec.~\ref{subsec:hyperparameters}, we give additional details on hyper-parameters and discuss them.

\subsection{Approximate partial policy evaluation}
\label{subsec:mstp_vs_mreg}

\begin{figure}
        \includegraphics[width=.8\linewidth]{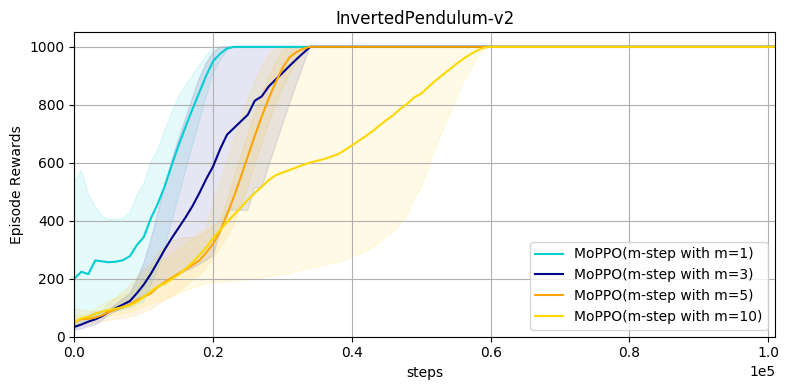}
        \hfill
        \includegraphics[width=.8\linewidth]{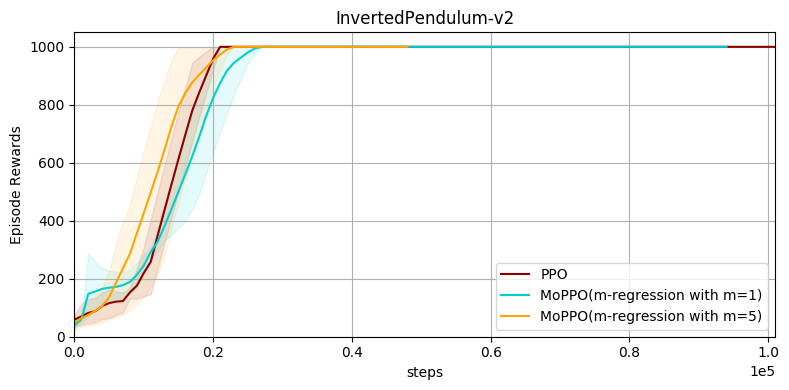}
\caption{Approximation of $(T_\pi)^m$ (inverted pendulum). Up, $m$-step rollouts. Down, $m$ regressions.\label{fig:Invm}}  
\end{figure}

\begin{figure}
        \includegraphics[width=.8\linewidth]{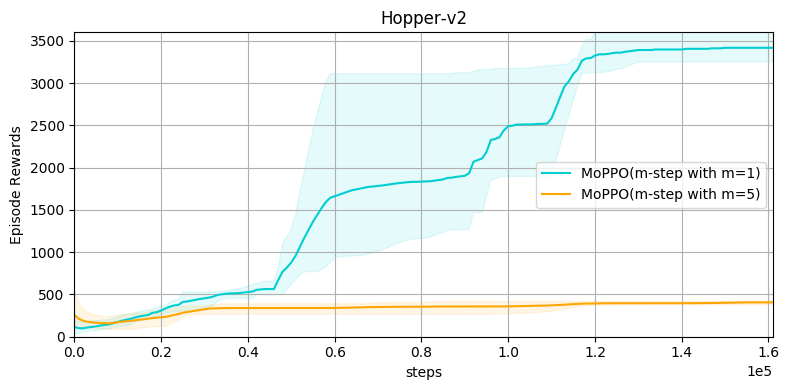}
        \hfill
        \includegraphics[width=.8\linewidth]{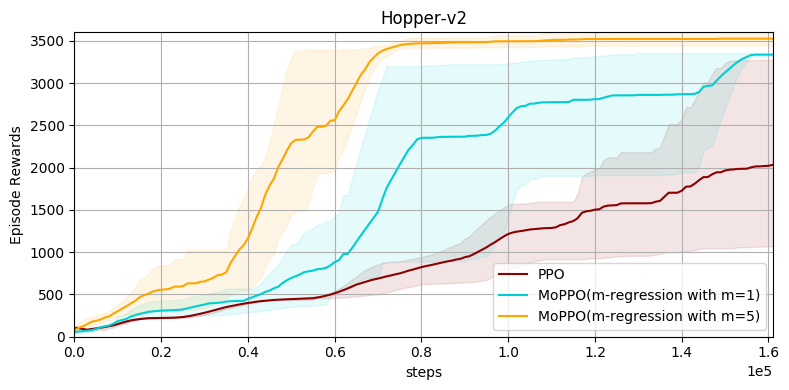}
    \caption{Approximation of $(T_\pi)^m$ (Hopper). Up, $m$-step rollouts. Down, $m$ regressions.\label{fig:Hopperm}}  
\end{figure}

As discussed in Sec.~\ref{subsec:mpi}, to approximate the operator $(T_\pi)^m$, one can either use $m$-steps rollouts (corrected for off-policy learning) or apply repeatedly $m$ times the approximate Bellman operator (or, said otherwise, solve $m$ regression problems). We have adopted the latter approach for MoPPO, but we also experimented the former. 

Rollouts off-policy correction is based on importance sampling, that can cause a huge variance (importance weights are ratio of probabilities, that can explose if probabilities are very different). To mitigate this effect, one can use the idea of Retrace~\citep{munos2016safe}, that consists in capping the importance weights at~1. We considered directly $m$-step rollouts corrected with Retrace.

Both approaches ($m$ regressions and $m$-step) work well on simple problems such as the InvertedPendulum as seen in Fig.~\ref{fig:Invm}. Yet, one can observe that for the regression, increasing $m$ helps, while for the rollout approach increasing $m$ degrades the performance. When applied to a mid-size problem such as Hopper (Fig.~\ref{fig:Hopperm}), PPO combined with (off-policy) $m$-step returns reaches much worse performance than PPO when $m>1$. Notice that for both figures, PPO (in red) corresponds to $m=\infty$ and is on-policy.

Our assumption is that MoPPO is too aggressive regarding the degree of ``off-policyness'' for an $m$-step return-based approach to work. On the contrary, performing $m$ successive regressions does not require any off-policy correction, and thus does not suffer from this variance problem. Also, it has been reported that Retrace might not be stable under function approximation~\citep{touati2017convergent}. Moreover, we notice that $m$-step returns are sometimes used without correction in an off-policy context, for example for Rainbow~\cite{hessel2017rainbow}. We hypothesize that it is because learning is slow enough (and $m$ is small enough too) so the transitions in the replay buffer are close to be on-policy.

These experiments suggest that doing $m$ regressions rather than $m$-step returns is beneficial, and that choosing $m>1$ is also beneficial (with $m=1$, both approaches are equivalent). As a consequence, we think that the general MoSoPI scheme could also be useful for other off-policy actor-critics (that roughly consider $m=1$ or corrected $m$-step returns).

\subsection{Effect of some parameters}
\label{subsec:effect_parameters}

First, we study the influence of $m$ (performance is evaluated as by~\citet{wu2017scalable}). With $m=1$, we have an approximate softened value iteration approach, and it get closer to policy iteration as $m$ increases (that could lead to an increase of computational budget).

\begin{figure}%
    \centering
    \begin{subfigure}[b]{0.8\linewidth}
        \includegraphics[width=\linewidth]{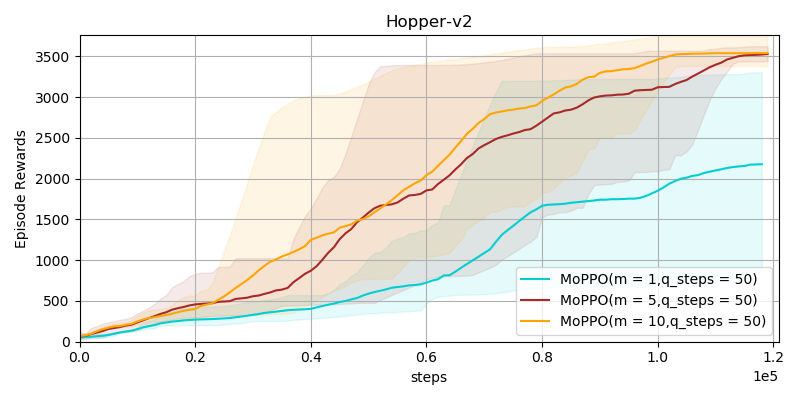}
        \caption{Varying $m$, proportional budget.}
        \label{fig:MfigUp}
    \end{subfigure}
    \hfill
    \begin{subfigure}[b]{0.8\linewidth}
        \includegraphics[width=\linewidth]{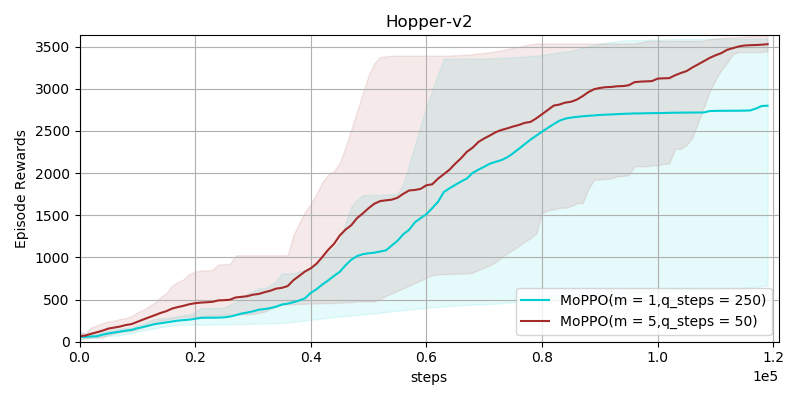}
        \caption{Varying $m$, fixed budget.}
        \label{fig:MfigUp2}
    \end{subfigure}
    \hfill
    \begin{subfigure}[b]{0.8\linewidth}
        \includegraphics[width=\linewidth]{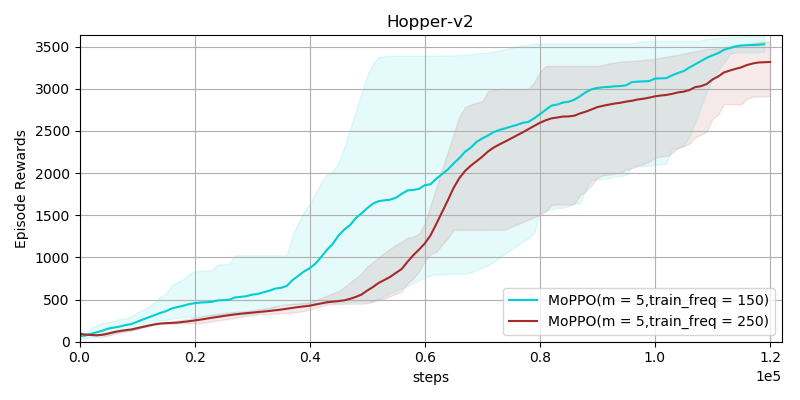}
        \caption{Varying updates' frequency.}%
        \label{fig:Mfig2}
    \end{subfigure}
    \caption{Influence of some parameters.}\label{fig:metaparams}
\end{figure}

In Fig.~\ref{fig:MfigUp}, we increase $m$ while keeping the budget of each regression fixed. That is, we process the same number of minibatches for each regression, here 50; the budget thus increases linearly with $m$. Going from $m=1$ to $m=5$ speeds-up learning and improves the final performance (it almost doubles). Going from $m=5$ to $m=10$ provides a smaller improvement. This suggests that we can gain something by solving repeatedly the regression problem corresponding to the Bellman operator, with a fixed policy.

This does not consume more samples, but comes with an increase of the computational budget. In Fig.~\ref{fig:MfigUp2}, we study the effect of increasing $m$ at a fixed budget (that is, keeping the number of minibatches being processed fixed for the whole set of $m$ regressions, to $m\times \text{q}\_\text{steps}=250$ in this case). In this example, increasing $m$, even with a fixed budget, still helps (but less than with an increase of the budget).

MoPPO %
decouples sample collection and learning, by updating both networks sequentially after every $\text{train}\_\text{freq}$ interactions with the environment. One can expect that if this parameter is too large, learning will be less sample efficient (as networks are updated less frequently), while if too small, learning could become more unstable. This is illustrated in Fig.~\ref{fig:Mfig2} for the Hopper task. With a more frequent update, learning is faster but variance is also higher.

\subsection{Comparative Results}
\label{subsec:comparison}

Here we compare MoPPO to PPO, which is quite natural. The only difference between both approaches is the way the advantage function is estimated. PPO is quite recent, and a standard baseline, but on-policy. Off-policy actor-critic algorithms is a fast evolving field, so we also compare our approach to SAC, that is also off-policy, and has state of the art performances.

We evaluate results using the approaches of~\citet{haarnoja2018soft} or of~\citet{wu2017scalable}.
The first one is representative of how learning progresses (current policy is evaluated every 1000 steps), while the other one is representative of the global efficiency of the algorithm (every 1000 steps, average of the best 10 policy evaluations computed so far).

In some environments, the performance of MoPPO degrades when the policy becomes too deterministic. We stop learning when this occurs, that explains why MoPPO curves sometime stop earlier\footnote{With the evaluation of~\citet{haarnoja2018soft}, continuing the curve would result in a possible degradation, while with the evaluation of~\citet{wu2017scalable}, it would lead to a saturation.}.

\begin{figure}%
    \centering
    \begin{subfigure}[b]{0.8\linewidth} 
        \includegraphics[width=\linewidth]{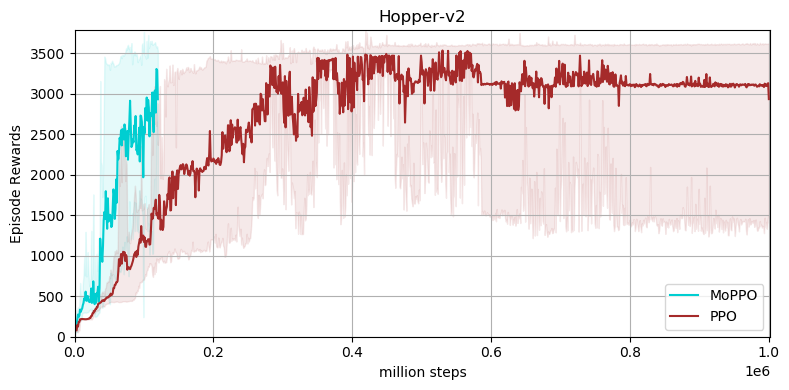}
    \end{subfigure}
    \hfill
    \begin{subfigure}[b]{0.8\linewidth}
        \includegraphics[width=\linewidth]{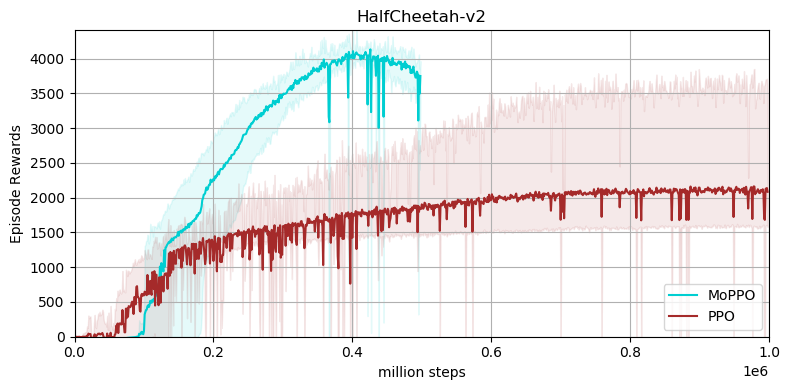}
    \end{subfigure}
    \hfill
    \begin{subfigure}[b]{0.8\linewidth}
        \includegraphics[width=\linewidth]{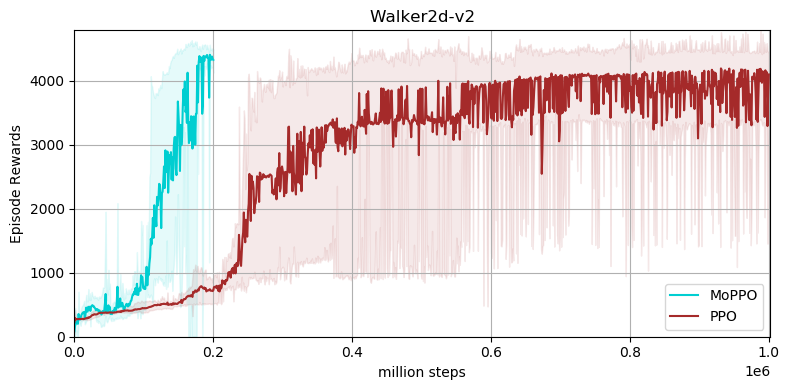}
    \end{subfigure}
    \hfill
    \begin{subfigure}[b]{0.8\linewidth}
        \includegraphics[width=\linewidth]{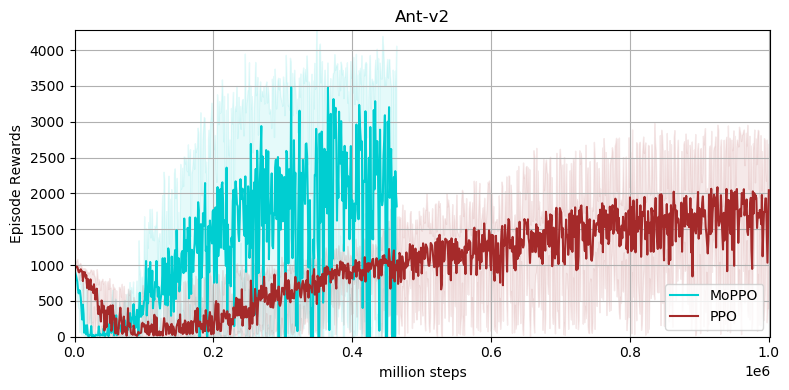}
    \end{subfigure}
    \hfill
    \begin{subfigure}[b]{0.8\linewidth}
        \includegraphics[width=\linewidth]{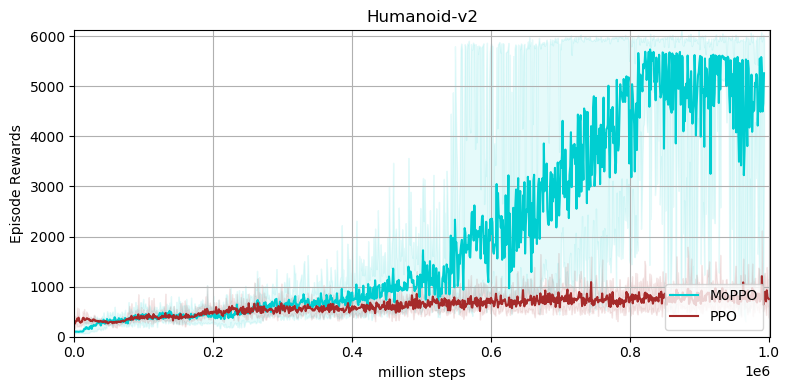}
    \end{subfigure}
    \caption{PPO vs MoPPO, policy is evaluated after every 1000 interactions with the environment.}
    \label{fig:Eval1}
\end{figure}

Fig.~\ref{fig:Eval1} shows the performances of PPO and MoPPO on five Mujoco tasks (see graph titles). We observe that MoPPO consistently learn competitive or better policies faster (up to 5 to 10 time faster, eg. Hopper or Walker). This was to be expected, as MoPPO is off-policy, while PPO is on-policy. However, we recall MoPPO to be a simple modification of PPO, this illustrates the fact that the general MoSoPI framework can be useful regarding sample efficiency. We can also observe that MoPPO can be less stable, its policy tends to become close to deterministic earlier, and learning can have more variance (eg. Ant, even if it still performs better in average than PPO).

\begin{figure}%
    \centering
    \begin{subfigure}[b]{0.8\linewidth} 
        \includegraphics[width=\linewidth]{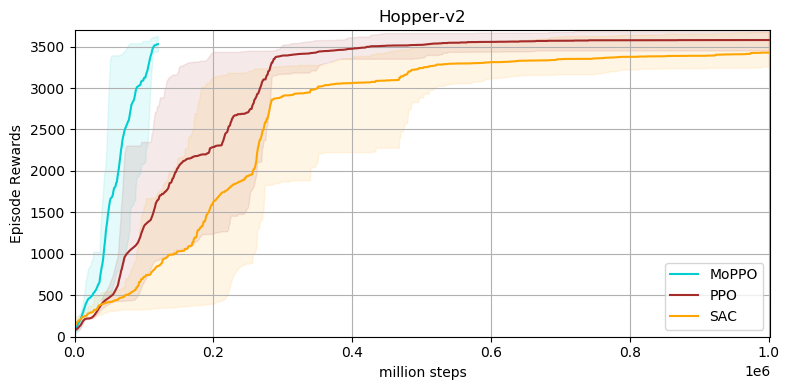}
    \end{subfigure}
    \hfill
    \begin{subfigure}[b]{0.8\linewidth}
        \includegraphics[width=\linewidth]{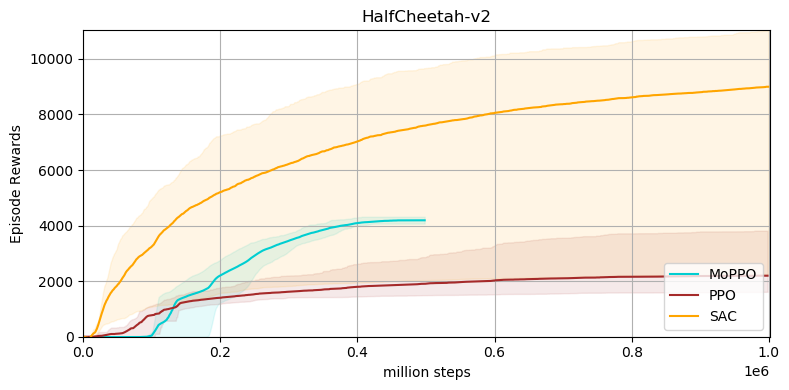}
    \end{subfigure}
    \hfill
    \begin{subfigure}[b]{0.8\linewidth}
        \includegraphics[width=\linewidth]{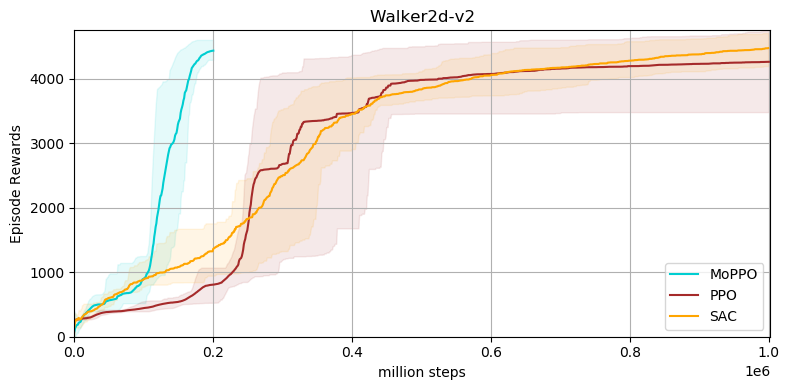}
    \end{subfigure}
    \hfill
    \begin{subfigure}[b]{0.8\linewidth}
        \includegraphics[width=\linewidth]{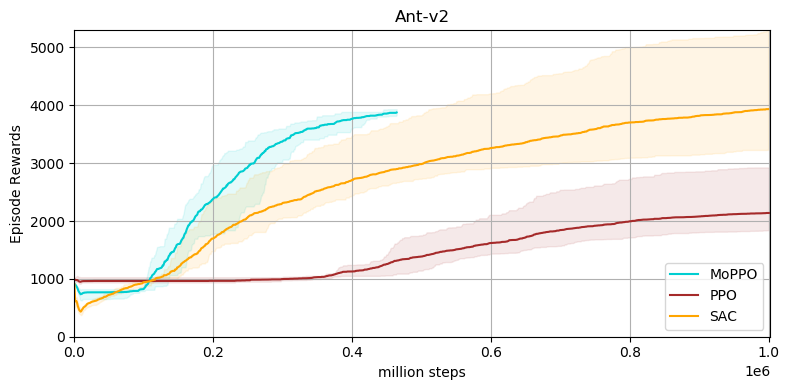}
    \end{subfigure}
    \hfill
    \begin{subfigure}[b]{0.8\linewidth}
        \includegraphics[width=\linewidth]{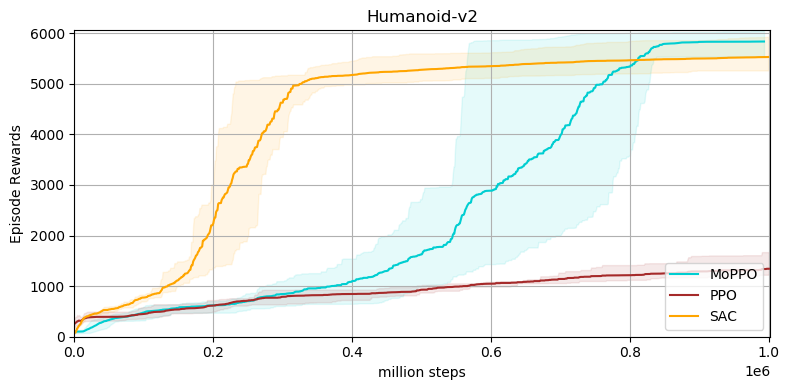}
    \end{subfigure}
    \caption{PPO vs SAC vs MoPPO. Results show the average of the past top ten evaluation runs after every 1000 interactions with the environment.}
    \label{fig:Eval10}
\end{figure}

Fig.~\ref{fig:Eval10} compares the average of the past top ten  policies for PPO, SAC and MoPPO. The comparison to PPO is as before. MoPPO performs as well as SAC in most environments, usually with much less samples. For example, in Walker it takes 5 times less sample to reach the same score as SAC. MoPPO is slower than SAC for Humanoid, but it reaches a better score (that SAC eventually reaches after 1.75 million interactions). It's only for HalfCheetah that SAC obtains clearly better results, and faster.

\subsection{Hyper-parameters}
\label{subsec:hyperparameters}

\begin{table*}[!htb]
\centering
\caption{Hyper-parameters for MoPPO for each environment.
\label{tab:hyper}}
\begin{center}
\begin{tabular}{llllll}
\toprule
~& \textbf{Hopper} & \textbf{HalfCheetah} & \textbf{Walker2d} & \textbf{Ant} & \textbf{Humanoid} \\
\midrule
$train\_freq$                          & 150             & 250                  & 150               & 250          & 1000              \tabularnewline
$m$                                    & 5               & 5                    & 5                 & 1            & 1                 \tabularnewline
$q\_steps$                             & 50              & 250                  & 50                & 50           & 500               \tabularnewline
$pol\_steps$                           & 500             & 500                  & 500               & 500          & 500               \tabularnewline
clip ratio                           & 0.005           & 0.005                & 0.005             & 0.0001       & 0.00005           \tabularnewline
buffer size                          & 20k             & 20k                  & 20k               & 20k          & 20k               \tabularnewline
batch size                           & 250             & 250                  & 250               & 250          & 250               \tabularnewline
normalized obs.              & Yes             & Yes                  & Yes               & Yes          & Yes               \tabularnewline
dual $Q$-Networks                      & No              & Yes                  & Yes               & Yes          & Yes               \tabularnewline
optimizer($Q$)                         & Adam(1e-3)      & Adam(1e-3)           & Adam(1e-3)        & Adam(1e-3)   & Adam(1e-3)        \tabularnewline
optimizer(Policy)                    & Adam(1e-4)      & Adam(1e-4)           & Adam(1e-4)        & Adam(1e-4)   & Adam(1e-4)       \tabularnewline
discount factor                      & 0.99            & 0.99                 & 0.99              & 0.99         & 0.99              \tabularnewline
gradient clipping                    & Yes             & Yes                  & Yes               & Yes          & Yes    
\end{tabular}
\end{center}
\end{table*}

Here, we provide additional details on considered hyper-parameters. For all environments, we used the state normalization provided by the OpenAI framework. The networks architectures are as follows.
For MoPPO and all experiments except Ant, the actor is a Gaussian policy with 2 hidden layers $(64,64)$ with $\tanh$ activation for each layer output and the critic is feedforward neural network with (400,300) using $\relu$ activations for each layer output. For the Ant environment, a larger policy is used, with 2 hidden layers of size $(400,300)$. The critic is the same.
PPO shares the same policy architecture and uses the same architecture for state value function as MoPPO's state-action value function. The only difference is the bigger input for MoPPO (state-action instead of state).
All other hyper-parameters are provided in Tab.~\ref{tab:hyper}.

Besides the $m$ and $\text{train}\_\text{freq}$ that we discussed in Sec.~\ref{subsec:effect_parameters}, there is indeed a few other parameters to set (but less strongly linked to the change of how the policy is evaluated). In our experiments, we observed that hyper-parameter setting is important for getting best results from the algorithm, due to variety of Mujoco tasks. For example, while Hopper has 11-dimensional state, Humanoid has 376-dimensional state (and is usually considered as a difficult problem).
MoPPO is an off-policy algorithm that uses a replay buffer. Best empirical results are achieved using a buffer size of 20k. Larger buffer sizes reduced the performance, maybe because larger buffers increase the degree of ``off-policyness'', other things being fixed (it would contain data from more different policies). Using smaller buffers tends to make learnt policies more greedy on a smaller set of actions, and training focuses on state-action pairs that are sampled by similar policies. 
The clip ratio of PPO is set to 0.2, its typical value, while MoPPO uses a much smaller clipping ratio ($5.10^{-3}$ to $5.10^{-5}$). We do so because more gradient descent steps are applied to the policy network (compared to PPO). The number of gradient descent steps applied to the $Q$-function, for each of the $m$ regressions, depends on the scale of the environment. In Hopper, we used 50-250 gradient steps, while large tasks required more gradient steps (reaching 1000 steps for Humanoid). We also not that if $m>1$ consistently provide good results, better results were obtained for $m=1$ on two tasks (Ant and Humanoid) within 1 million environment steps. On Humanoid task, $m=10$ can achieve around $10\%$ higher results but requires more environment steps (additional 120k environment step) and computations (10 times more).

\section{Conclusion}

In this paper, we proposed MoSoPI, a general framework that mixes the general idea of softened greediness, initiated by~\citet{kakade2002approximately}, with the partial evaluation approach of MPI~\citep{puterman1978modified}. As a proof of concept, we introduced MoPPO, a modification of PPO that changes the way the advantage is estimated, and allows for off-policy learning. In our experiments, MoPPO consistently learns faster and provides better policies than PPO. This simple modification is also competitive with SAC, and often (but not always) performs better and/or faster.

We would like again to highlight that our core contribution is a new tool in the deep RL toolbox.
MoPPO is just a proof of concept, and the general framework of MoSoPI can be used to derive other algorithms. For example, if MoPPO learns much faster than PPO, we clearly pay this by less stability, as discussed earlier. As such, it is efficient for learning a good policy from a small amount of samples (as we can keep the best policy computed so far), but would not be an ideal solution for continuous learning.

We envision to use different kinds of softened greediness to help stabilize learning. More specifically, we plan to take inspiration from~\citet{abdolmaleki2018relative} or~\citet{haarnoja2018softb}, who basically stabilize learning by controlling better the entropy of the policy and/or how it evolves.
Another approach could be to select in a smarter way what experience to keep in the replay buffer~\cite{de2018experience}.
One could also envision to reload the last best policy after sample collection, if the last one degrades too much performance. This is possible, as we disentangle learning and sample collection, and if one does this, the next policy would be learnt from a more diverse replay buffer (containing samples collected by the bad policy), and could lead to a better policy (or at least to a different less good policy).
We think that combining these ideas with the partial policy evaluation scheme proposed here could further improve sample efficiency.

\bibliographystyle{ACM-Reference-Format}  %
\bibliography{sample-bibliography}  %

\end{document}